\documentclass[journal,11pt,draftclsnofoot,onecolumn]{IEEEtran}

\ifCLASSINFOpdf

\else

\fi

\usepackage{booktabs}
\usepackage[justification=centering]{caption}
\usepackage{latexsym,amsfonts,amssymb,graphicx,graphics,tikz}
\usepackage{amssymb}
\usepackage{mathrsfs}
\usepackage{pst-poly}  
\usepackage{cite}
\usepackage[numbers,sort&compress]{natbib}
\usepackage{multicol}
\usepackage{hhline}
\usepackage{amsmath}
\usepackage{subcaption}
\usepackage{graphicx}
\usepackage{array}
\usepackage{mdwmath}
\usepackage{mdwtab}
\usepackage{algorithmic}
\usepackage{breqn}
\usepackage{eqparbox}
\usepackage{fancyhdr}
\usepackage{balance}
\usepackage[ruled,linesnumbered,boxed]{algorithm2e}
\usepackage{multirow}                
\usepackage{amsmath}
\usepackage{geometry}
\usepackage{makecell}
\usepackage{xcolor}
\usepackage{caption}
\usepackage{chngcntr}
\counterwithout{table}{section}
\usepackage{graphics}
\usepackage{epsfig}

\def\theequation{\arabic{equation}}

\renewcommand{\tablename}{Table}

\begin{document}
\title{\bf A Survey on Graph Classification and Link Prediction based on GNN}

{\baselineskip 12pt \vskip0.05cm
\author{  Xingyu Liu\quad Juan Chen\quad Quan Wen\\
\small School of Computer Science and Engineering \\
\small University of Electronic Science and Technology of China \\
\small Chengdu, Sichuan, 611730, P.R. China\\
}}
\date{}
\maketitle

{\bf\noindent Abstract:}\ Traditional convolutional neural networks are limited to handling Euclidean space data, overlooking the vast realm of real-life scenarios represented as graph data, including transportation networks, social networks, and reference networks. The pivotal step in transferring convolutional neural networks to graph data analysis and processing lies in the construction of graph convolutional operators and graph pooling operators. This comprehensive review article delves into the world of graph convolutional neural networks. Firstly, it elaborates on the fundamentals of graph convolutional neural networks. Subsequently, it elucidates the graph neural network models based on attention mechanisms and autoencoders, summarizing their application in node classification, graph classification, and link prediction along with the associated datasets. 
\vskip6pt

{\noindent{\bf Keywords:}  Graph convolutional neural network, Node classification, Link prediction.

\section{\it\bf Introduction }

The characteristic of deep learning is the accumulation of multiple layers of neural networks, resulting in better learning representation ability. The rapid development of convolutional neural networks (CNN) has taken deep learning to a new level\cite{l1,l2}. The translation invariance, locality, and combinatorial properties of CNN make it naturally suitable for tasks such as processing Euclidean structured data such as images\cite{l3,l4}, At the same time, it can also be applied to various other fields of machine learning\cite{l5,l6,l7}. The success of deep learning partly stems from the ability to extract effective data representations from Euclidean data for efficient processing. Another reason is that thanks to the rapid development of GPUs, computers have powerful computing and storage capabilities, It can train and learn deep learning models in large-scale data sets, which makes deep learning perform well in natural language processing\cite{l8}, machine vision\cite{l9}, recommendation systems\cite{l10} and other fields

However, existing neural networks can only process conventional Euclidean structured data. As shown in Figure. 1(a), Euclidean data structures are characterized by fixed arrangement rules and orders of nodes, such as 2D grids and 1D sequences. Currently, more and more practical application problems must consider non Euclidean data, such as  Figure. 1(b), where nodes in non Euclidean data structures do not have fixed arrangement rules and orders, This makes it difficult to directly transfer traditional deep learning models to tasks dealing with non Euclidean structured data. If CNN is directly applied to it, it is difficult to define convolutional kernels in non Euclidean data due to the unfixed number and arrangement order of neighboring nodes in the non Euclidean data center, which does not meet translation invariance. Research work on graph neural networks (GNNs), At the beginning, it was about how to fix the number of neighboring nodes and how to sort and expand them, such as the PATCHY-SAN\cite{l11}, LGCN\cite{l12}, DCNN\cite{l13} methods. After completing the above two tasks, non Euclidean structured data is transformed into Euclidean structured data, which can then be processed using CNN. A graph is a typical non Euclidean data with points and edges, In practice, various non Euclidean data problems can be abstracted into graph structures. For example, in transportation systems, graph based learning models can effectively predict road condition information\cite{l14}. In computer vision, the interaction between humans and objects can be viewed as a graph structure, which can be effectively recognized\cite{l15}.

\begin{figure}[htbp]
	\centering
	\begin{subfigure}{0.35\linewidth}
		\centering
		\includegraphics[width=1.1\linewidth]{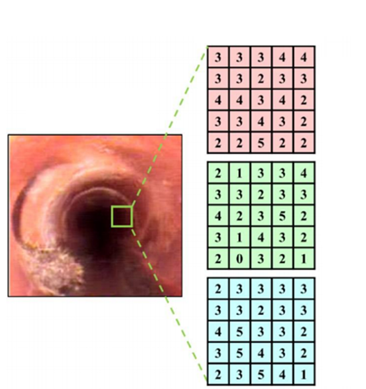}
		\caption{RGB value of an image (Euclidean)}
		\label{p1-1}
	\end{subfigure}
	\hfill
	\begin{subfigure}{0.4\linewidth}
		\centering
		\includegraphics[width=0.9\linewidth]{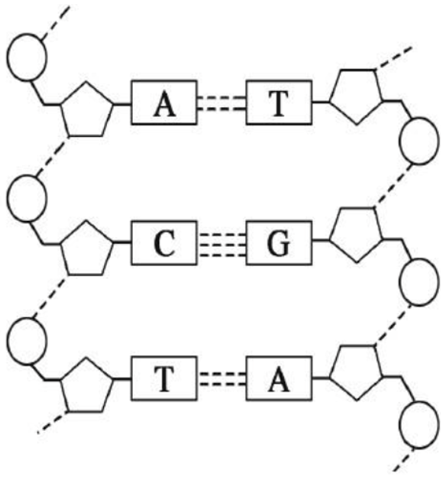}
		\caption{DNA molecular structure (non-Euclidean)}
		\label{p1-2}
	\end{subfigure}
	\caption{Euclidean and non-Euclidean data structure}
	\label{p1}
\end{figure}

Recently, some scholars have reviewed graph neural networks and their branches of graph convolutional neural networks\cite{l16,l17,l18}. The difference in this article is that it focuses on introducing the methods and models of graph neural networks in node classification and link prediction of citation networks. In citation networks, a typical classification task is to provide the content information and citation relationships between each article, and to classify each article into the corresponding domain. For example, in a semi supervised classification scenario of nodes, the attribute information of nodes includes the title or abstract information of the article, as well as the relationships referenced between nodes to form network information. Given a small amount of virtual data tables, the domain to which each node belongs in the network is divided through deep learning. In this task, GCN effectively modeled the node text attributes and reference network structure, achieving great success. Compared to directly using content information (such as MLP), using only structural information (such as DeepWalk\cite{l19}) and traditional semi supervised node classification methods on graphs, such as Planetoid \cite{l20}, traditional methods have much lower classification accuracy than graph convolutional neural network algorithms represented by GCN. Among them, the Graph Attention Network (GAT)\cite{l21} performs better than the Planetoid model in classic citation network datasets. Therefore, this task is often seen as a benchmark task to measure the effectiveness of a graph convolutional neural network model. GCN\cite{l22}, GAT\cite{l21}, and GWNN\cite{l23} all used citation network classification tasks to verify the effectiveness of the model.

\section{\it\bf Graph Neural Network}
\subsection{Graph Structure Class}
\subsubsection{Edge Information Graph}
In recent years, the concept of edge information graph has gained considerable attention in the field of graph theory. An edge information graph is defined as a graph structure in which different edges possess distinct structural characteristics. These characteristics may include the weight, direction, and heterogeneous relationships between nodes.

For example, consider the complex structure of a social network graph. The relationships between nodes within this graph may take on a variety of forms, ranging from unidirectional "follow" relationships to bidirectional "friendship" relationships. Due to the complexity of such relationships, they can not be adequately represented by simple weight constraints. This highlights the importance of considering the full range of structural edge information in the analysis of graphs with complex relationships.

\subsubsection{Spatio-Temporal Graph}
A Spatio-Temporal graph is a type of property graph.Its characteristic is that the characteristic matrix $X$ in the high-dimensional feature space $f^*$ will change with time. This structure is represented as $G^* = (V, E, A, X)$, where V, E, and A denote the vertices, edges, and adjacency matrix, respectively. With the introduction of time series, graph structure can effectively manage tasks that require handling of dynamic and temporal relationship types. Yan et al.\cite{liu4} presented a method for skeleton motion detection based on Spatio-Temporal graph convolutional neural networks.

\subsection{Convolution Graph Neural Network}
Graph convolutional neural networks can be divided into two categories in terms of feature space: frequency domain and spatial domain. A graph convolutional neural network maps the data $G = (V, E)$ of the original graph structure to a new feature space$f^G \rightarrow f^*$. Taking a single-layer forward propagation graph convolutional neural network as an example, the features of the layer $i$ neural network are denoted by $w_i$. In computing each node $v_i$ in the graph structure, the output $H^{l+1}$ of each layer of the neural network can be expressed by the nonlinear function $f(\cdot,\cdot)$, where $A$ is the feature adjacency matrix. The graph convolutional neural network structure is implemented by the nonlinear activation function ReLU = $\sigma(\cdot)$ with the following layered propagation rule:$f(H^l,A) = \sigma(\hat{D}^{-1/2}\hat{A}\hat{D}^{-1/2}H^lW^l)$ where $\hat{A} = A+I$ denotes the adjacency matrix of the graph structure $G = (V,E)$, $I$ denotes the identity matrix, $\hat{D} = \sum{\hat{A}_{ij}}$ denotes the diagonal matrix, and $W_l$ denotes the weight matrix of the layer $l$ of the convolutional neural network . Through the hierarchical propagation rules, the graph convolution neural network introduces the local parameter sharing characteristics of the convolution neural network into the graph structure, so that the breadth of the sensing area of each node will be greatly improved with the increase of the number of propagation layers, so as to obtain more information from the neighboring nodes.Based on the existing GNN structure, a general GNN structure flowchart can be and summarized, as shown in Figure. 2.

\begin{figure}[htbp]
	\centering
	\includegraphics[width=0.6\linewidth]{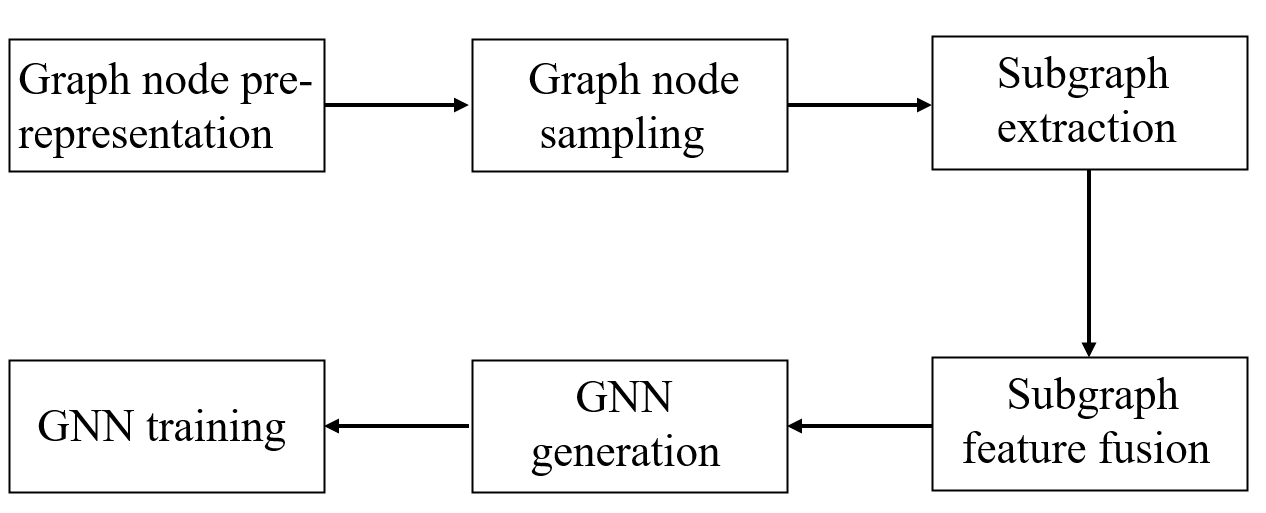}
	\caption{General structure of graph neural networks}
	\label{p0}
\end{figure}

\subsection{Spatio-Temporal Graph Neural Network}

As an attribute graph network, The Spatio-Temporal graph neural network introduces the characteristics of time series. It can simultaneously obtain the characteristic information of time and space domains in the graph structure, and the characteristics of each node will change with time. We mainly discusses the Spatio-Temporal graph neural network structure that uses graph convolution to extract spatial feature dependence in the spatial domain. It is mainly divided into three time-domain feature acquisition methods: traditional convolution network, gated loop network and graph convolution network. Figure. 3 shows the network structure comparison between graph convolution neural network and Spatio-Temporal graph neural network (taking 1D-CNN+GCN structure as an example). The two network structures are constructed on the basis of graph convolution computing unit, where $\varphi$ Is the element distance between matrix $Z$ and $Z^T$, and MLP full connection represents multilayer perceptron full connection neural network.

\begin{figure}[!ht]
 \centering
 \resizebox*{4in}{!}{
 \includegraphics{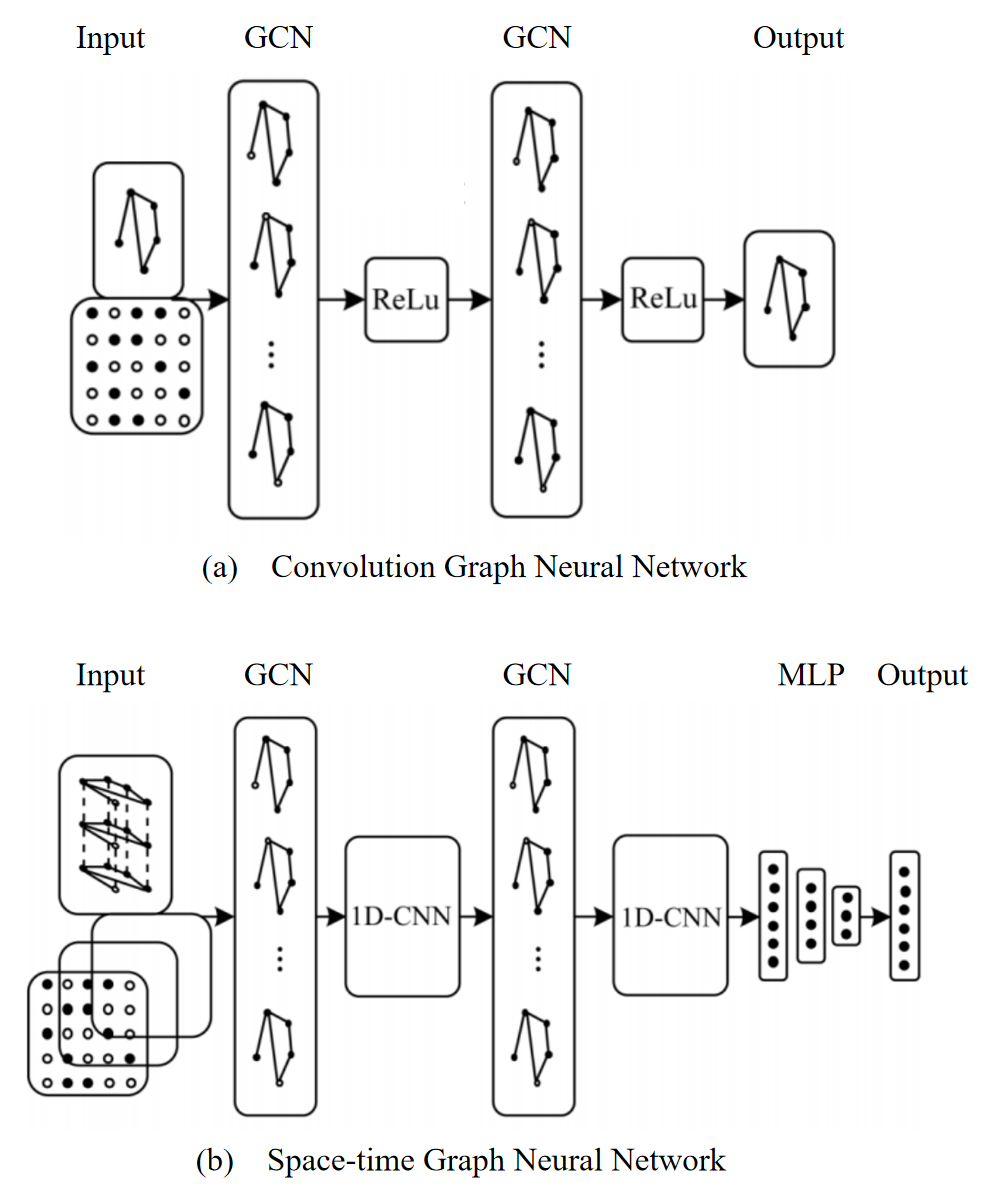}}
 \caption{The spatial temporal graph of a skeleton sequence.}
 \label{lxy1}
\end{figure}

\section{\it\bf Graph Neural Network Based on Attention Implementation}
The attention mechanism has shown strong capabilities in processing sequential tasks\cite{l24}, such as in machine reading and learning sentence representation tasks. Its powerful advantage lies in allowing variable input sizes, and then utilizing the attention mechanism to only focus on the most important parts before making decisions. Some studies have found that the attention mechanism can improve convolutional methods, allowing for the construction of a powerful model, In dealing with some tasks, better performance can be achieved. Therefore, reference\cite{l21} introduced attention mechanism into the process of neighbor node aggregation in graph neural networks and proposed graph attention networks (GAT). In the traditional GNN framework, attention layers were added to learn the different weights of each neighbor node, Treat them differently. In the process of aggregating neighboring nodes, only focus on the nodes with larger effects, while ignoring some nodes with smaller effects. The core idea of GAT is to use neural networks to learn the weights of each neighboring node, and then use neighboring nodes with different weights to update the representation of the central node. Figure. 4 is a schematic diagram of the GAT layer structure. Figure. 4(a) shows the calculation of weights between node $i$ and node $j$, Figure. 4(b) shows a node using a multi head attention mechanism in its neighborhood to update its own representation. The attention factor of node $j$ relative to node $i$ is solved as:

\begin{figure}[htbp]
	\centering
	\begin{subfigure}{0.4\linewidth}
		\centering
		\includegraphics[width=1\linewidth]{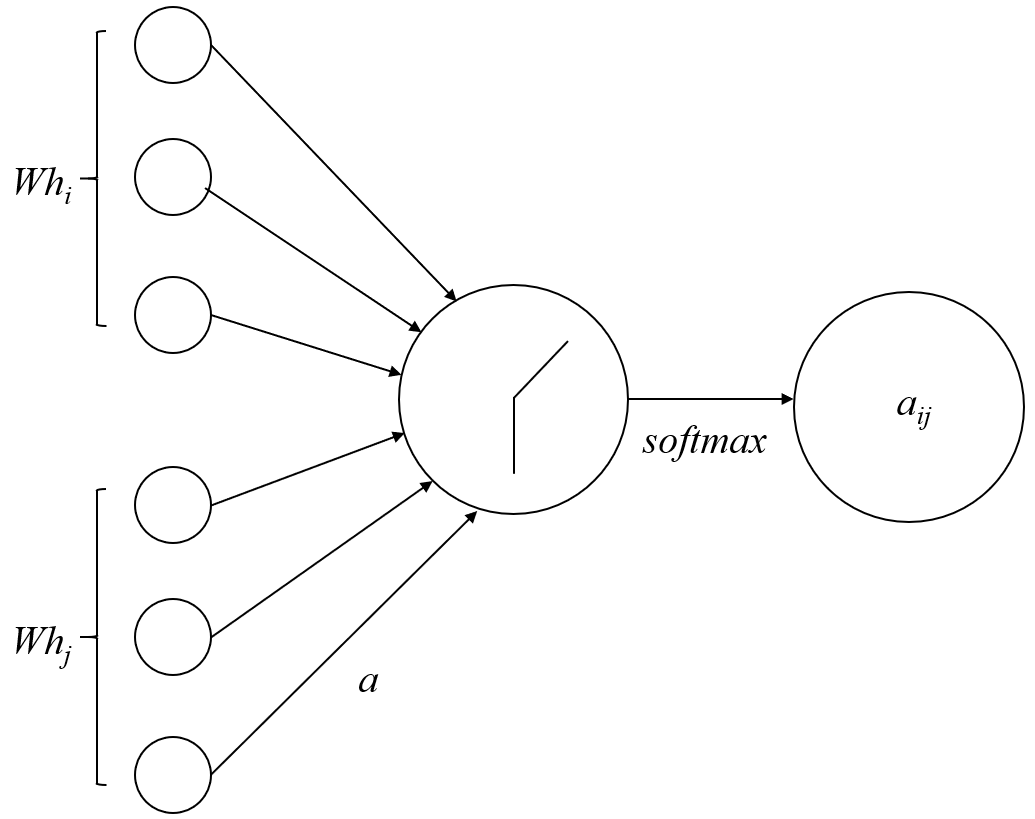}
		\caption{The attention mechanism $a(\boldsymbol{W}\vec{h_i},\boldsymbol{W}\vec{h_j})$}
		\label{p1-1}
	\end{subfigure}
	\begin{subfigure}{0.55\linewidth}
		\centering
		\includegraphics[width=1.2\linewidth]{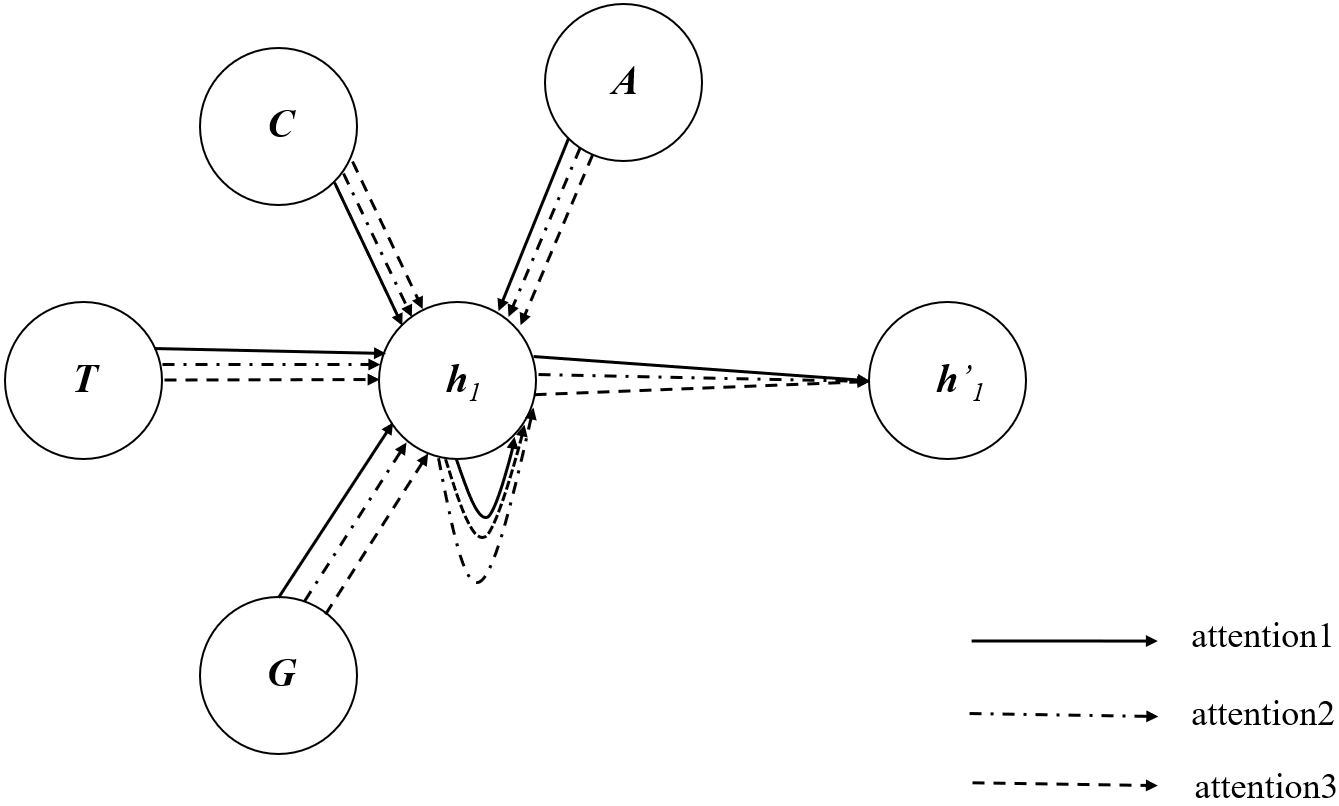}
		\caption{An illustration of multi-head attention }
		\label{p1-2}
	\end{subfigure}
	\caption{Structure of GAT layer}
	\label{p1}
\end{figure}

\begin{equation}
a_{i j}=\frac{\exp \left(L e a k y \operatorname{ReLU}\left(\boldsymbol{\alpha}^{\mathrm{T}}\left[\boldsymbol{W h}_i \| \boldsymbol{W h}_j\right]\right)\right)}{\sum_{k \in N_i} \exp \left(L e a k y \operatorname{ReLU}\left(\boldsymbol{\alpha}^{\mathrm{T}}\left[\boldsymbol{W h} \| \boldsymbol{W h}_j\right]\right)\right)}
\end{equation}
where $a_ij$ represents the attention factor of node $j$ relative to node $i$, \boldmath{$W$} is a affine transformation for dimension reduction, $\boldmath{\alpha}^T$ represents the weight vector parameter,$||$ represents the vector splicing operation, and $LeakyReLU(x^') = \left\{\begin{array}{l}x^{\prime}, x^{\prime}>0 \\ \lambda x^{\prime}, x^{\prime} \leq 0\end{array}\right.$ is the leak correction linear unit. Then, with the nonlinear activation function $\delta$, the learned attention factor $a_ij$ can be used to update the central node $i$:

\begin{equation}
\boldsymbol{h}_i^{\prime}=\delta\left(\sum_{j \in N_i} a_{i j}^k \boldsymbol{W}^k \boldsymbol{h}_j\right)
\end{equation}

In order to make the model more stable, the author also applied a multi head attention mechanism. Instead of using only one function to calculate attention factors, $K$ different functions were set to jointly calculate attention factors. The results of each function can obtain a set of attention parameters, and can also provide a set of parameters for the weighted sum of the next layer. In each convolutional layer, $K$ different attention mechanisms do not affect each other, Work independently. Finally, concatenate or average the results obtained from each attention mechanism to obtain the final result. If $K$ different attention mechanisms are calculated simultaneously, we can obtain:

\begin{equation}
\boldsymbol{h}_i^{\prime}=\|_{k=1}^K \delta\left(\sum_{j \in N_i} a_{i j}^k \boldsymbol{W}^k \boldsymbol{h}_j\right)
\end{equation}
$||$ represents the concatenation operation, and $a_{ij}^k$ is the attention factor obtained by the $k$-th attention parameter function. For the last convolutional layer, if the multi head attention mechanism is used for solving, the average method should be used to solve:

\begin{equation}
\boldsymbol{h}_i^{\prime}=\delta\left(\frac{1}{K} \sum_{k=1}^K \sum_{j \in N_i} a_{i j}^k \boldsymbol{W}^k \boldsymbol{h}_j\right)
\end{equation}

Reference \cite{l25} also introduced the multi head attention mechanism into the aggregation process of neighboring nodes, proposing gated attention networks (GAAN). However, unlike GAT, which uses averaging or concatenation to determine the final attention factor, GAAN believes that although using the multi head attention mechanism can gather information from multiple neighboring nodes of the central node, not every head of attention mechanism has the same contribution, A certain head of attention may capture useless information. Therefore, GAAN assigns different weights to each attention mechanism in multi head attention to aggregate neighboring node information and complete the update of the central node. Therefore, GAAN first calculates an additional soft gate between 0 (low importance) and 1 (high importance), assigning different weights to each head of attention. Then, combined with the multi head attention aggregator, You can obtain a gated attention aggregator:

\begin{equation}
y_i=F C_{\theta_0}\left(\tilde{\boldsymbol{x}}_i \oplus \prod_{k=1}^K\left(g_i^{(k)} \sum_{j \in N} w_{i, j}^{(k)} F C_{\theta_v^{(k)}}^h\left(z_j\right)\right)\right)
\end{equation}
\begin{equation}
\boldsymbol{g}_i=\left(g_i^{(1)}, g_i^{(2)}, \cdots, g_i^{(K)}\right)=\psi_g\left(\tilde{\boldsymbol{x}}_i, z_{N_i}\right)
\end{equation}
where $FC_{\theta_{0}}(\cdot)$ means that the activation function is not applied after the linear transformation, $\bigoplus$ is the connection operation, $K$ is the number of attention mechanisms, and $w_{i,j}^{(k)}$ is the $k$-th attention weight between node $i$ and $j$, $\theta_v^{(k)}$ is the parameter of the $k$-th header used to query the vector. $g_i^{(k)}$ is the threshold value of the $k$-th header of node $i$, Apply convolutional network $\Psi_g$ and take the center node feature $\widetilde{x}_i$ and neighbor node feature $z_{N_i}$ to calculate the $g_i$. Convolution network $\Psi_g$.  The convolutional network $\Psi_g$ can be designed according to its actual needs, and literature\cite{l26} adopts average pooling and maximum pooling for construction:

\begin{equation}
\begin{gathered}\boldsymbol{g}_i=F C_{\theta_g}^\delta\left(\tilde{\boldsymbol{x}}_i \oplus \max _{j \in N_i}\left(\left\{F C_{\theta_m}\left(\boldsymbol{z}_j\right)\right\}\right) \oplus\right. \left.\frac{1}{\left|N_i\right|} \sum_{j \in N_i} z_j\right),\end{gathered}
\end{equation}
where $\theta_m$ represents mapping the feature vectors of neighboring nodes to the dimension $d_m$, $\theta_g$ represents mapping the concatenated feature vectors to the $k$-th gate. Finally, the author of reference\cite{l26} constructed a gated recursive unit using GGAN and successfully applied it to traffic speed prediction problems.

In reference\cite{l25}, it was proposed that although GAT has achieved good results in multiple tasks, there is still a lack of clear understanding of its discriminative ability. Therefore, the author of this paper conducted a theoretical analysis of the representation characteristics of graph neural networks using attention mechanisms as aggregators, and analyzed that such graph neural networks are always unable to distinguish all situations with different structures. The results show that, The existing attention based aggregators cannot preserve the cardinality of multiple sets of node feature vectors during aggregation, which limits their discriminative ability. The proposed method modifies the cardinality and can be applied to any type of attention mechanism.Zhang et al\cite{l27} developed a self attention graph neural network (SAGNN) based on attention mechanism for hypergraphs. SAGNN can handle different types of hypergraphs and is suitable for various learning tasks and isomorphic and heterogeneous hypergraphs with variables. This method can improve or match the latest performance of hypergraph learning, solving the shortcomings of previous methods, For example, it is impossible to predict the hyperedges of non-uniform heterogeneous hypergraphs. U2GNN\cite{l28} proposed a novel graph embedding model by introducing a universal self attention network, which can learn low dimensional embedding vectors that can be used for graph classification. In implementation, U2GNN first uses attention layers for calculation, Then, a recursive transformation is performed to iteratively remember the weight size of the vector representation of each node and its neighboring nodes in each iteration, and the final output sum is the final embedded representation of the entire graph. This method can solve the weaknesses in existing models, To generate reasonable node embedding vectors, the above models apply attention mechanism to spatial domain graph neural networks. In order to better utilize the local and global structural information of the graph, reference\cite{l29}first attempted to transfer attention mechanism from spatial domain to spectral domain, proposing spectral graph attention network (SpGAT). In SpGAT, graph wavelets are selected as spectral bases, And decompose it into low-frequency and high-frequency components based on indicators. Then, construct two different convolutional kernels based on low-frequency and high-frequency components, and apply attention mechanisms to these two kernels to capture their importance. By introducing different trainable attention weights to low-frequency and high-frequency components, local and global information in the graph can be effectively captured, And compared to the spatial domain, the attention spGAT greatly reduces learning parameters, thereby improving the performance of GNN. In order to better understand the application of attention mechanisms in graph neural networks and identify the factors that affect attention mechanisms, a series of experiments and models were designed in reference \cite{l30} to conduct in-depth research and analysis. Firstly, the graph isomorphism network (GIN) model\cite{l31} was used to conduct experiments on the dataset, but it was found that its performance was very poor, And it is difficult to learn attention subgraph networks. Therefore, the author combined GIN and ChebyNet networks to propose a ChebyGIN network model, and added attention factors to form an attention model. A weakly supervised training method was adopted to improve the performance of the model, Experiments were conducted on the models in color counting and triangle counting tasks, and four conclusions were drawn: 
\subsubsection{}
The main contribution of the attention mechanism in graph neural networks to node attention is that it can be extended to more complex or noisy graphs, which can transform a model that cannot be generalized into a very robust model; 
\subsubsection{}
The factors that affect the performance of attention mechanism in GNN include the initialization of attention model, the selection of GNN model, attention mechanism and the hyperparameter of GNN model; 
\subsubsection{}
Weak supervised training methods can improve the performance of attention mechanisms in GNN models; 
\subsubsection{}
The attention mechanism can make GNN more robust to larger and noisy graphs. 

We summarize the attention based graph convolutional neural network model mentioned above in Table 1:

\begin{table}
\caption{\tablename~{Attention-Based Graph Neural Network Models}}
\centering
\setlength{\tabcolsep}{10pt}
\renewcommand{\arraystretch}{1.5}
\resizebox{\textwidth}{!}{
\begin{tabular}{ccccp{6cm}}
\hline
Model & Year & Dataset & Application & \thead{Disadvantage} \\
\hline
GAT\cite{l21} & 2018 & \makecell{Citeseer, Cora,\\Pubmed, PPI} & Node classification & Limited scalability and local focus \\
GaAN\cite{l26} & 2018 & \makecell{PPI, Reddit,\\METR-LA} & Node classification & Hyperparameter sensitivity and limited long-range capture \\
SAGNN\cite{l27} & 2020 & \makecell{GPS, MovieLens,\\Drug, Wordnet} & Classification & Limited generalization and difficulty with high-order structures \\
U2GNN\cite{l28} & 2019 & \makecell{MUTAG, D\&D,\\Pubmed} & Graph classification & High computational complexity and potential over-smoothing \\
SpGAT\cite{l29} & 2020 & \makecell{Citeseer, Cora,\\Pubmed} & Node classification & Scalability issues and sensitivity to sparsity \\
ChebyCIN\cite{l30} & 2019 & \makecell{Olors, Triangles,\\D\&D} & Classification generation & High computational complexity and limited long-range capture \\
\hline
\end{tabular}
}
\end{table}

\section{\it\bf Graph Neural Network Based on Autoencoder Implementation}
In the unsupervised learning task, the autoencoder (AE) and its variants play a very important role. It realizes implicit representation learning with the help of neural network model, and has strong data feature extraction ability. AE realizes effective representation learning of input data through encoder and decoder, and the dimension of implicit representation learned can be far less than the dimension of input data, The purpose of dimensionality reduction is achieved. AE is currently the preferred deep learning technology for implicit representation learning. When we input raw data with certain connections $(\boldmath{x_1},\boldmath{x_2},\cdots,\boldmath{x_n})$ into AE for reconstruction learning, we can complete the task of feature extraction. The application scenarios of autoencoders are very wide, and they are often used in tasks such as data denoising, image reconstruction, and anomaly detection. In addition, When AE is used to generate data similar to training data, it is called a generative model. Due to the above advantages of AE, some scholars have applied AE and its variant models to graph neural networks. Reference\cite{l32} first proposed a variational graph autoencoder (VGAE) model based on variational autoencoder (VAE), Apply VAE to the processing of graph structured data. VGAE uses hidden variables to learn interpretable hidden representations of undirected graphs, and implements this model using a graph convolutional network encoder and a simple inner product decoder. In this model, the encoder is implemented using a 2-layer GCN:
\begin{equation}
q(\boldsymbol{H} \mid \boldsymbol{I}, \boldsymbol{A})=\prod_{i=1}^N q\left(\boldsymbol{h}_i \mid \boldsymbol{I}, \boldsymbol{A}\right)
\end{equation}
where $q\left(\boldsymbol{h}_i \mid \boldsymbol{I}, \boldsymbol{A}\right)=N\left(\boldsymbol{h}_i \mid \mu_i, \operatorname{diag}\left(\delta_i^2\right)\right)$, the average matrix of nodes is $\mu=GCN_{\mu}(I,A)$, and the variance of nodes is $log\sigma=GCN_{\sigma}(I,A)$. The GCN of layer 2 is:
\begin{equation}
    GCN(I,A)=\widetilde{A}ReLU(\widetilde{A}XW_0)W_1
\end{equation}
where, $\widetilde{\boldsymbol{A}} \widetilde{\boldsymbol{D}}=\boldsymbol{D}^{-\frac{1}{2}} \boldsymbol{A} \boldsymbol{D}^{-\frac{1}{2}}$is the adjacency matrix of the symmetric specification. The generative model used to reconstruct the graph is calculated by using the inner product of implicit variables:$p(\boldsymbol{A} \mid \boldsymbol{I})=\prod_{i=1}^N \prod_{j=1}^N p\left(A_{i j} \mid \boldsymbol{h}_i, \boldsymbol{h}_j\right)$, where $p(A_{ij}=1|h_i,h_j)=\sigma(h^T_i,h_j)$, $A_ij$ are the elements of matrix $A$. Finally, the loss function is defined as:
\begin{equation}
  \begin{aligned} & L=E_{q(\boldsymbol{H} \mid \boldsymbol{I}, \boldsymbol{A})}[\log p(\boldsymbol{A} \mid \boldsymbol{H})]- KL[q(\boldsymbol{H} \mid \boldsymbol{I}, \boldsymbol{A}) \| P(\boldsymbol{H})]\end{aligned}  
\end{equation}
where, the first item on the right side of the equation represents the cross entropy function, and the second item represents the $KL$ distance between the graph generated by the decoder and the input graph.

Most of the existing network embedding methods represent each node by a point in the low dimensional vector space. Thus, the formation of the entire network structure is deterministic. However, in reality, the network is full of uncertainty in the process of formation and evolution, which makes these methods have some drawbacks. In view of the above drawbacks, Reference\cite{l33} proposed a deep variational network embedding (DVNE) method for embedding in Wasserstein space. Due to the fact that Gaussian distributions essentially represent uncertainty properties, DVNE utilizes a deep variational model to learn Gaussian embeddings for each node in Wasserstein space, rather than using a point vector to represent nodes. This allows for the learning of Gaussian embeddings for each node in Wasserstein space while maintaining network structure, Modeling the uncertainty of nodes. In the DVNE method, the second Wasserstein distance ($W_2$) is used to measure the similarity between distributions. A deep variational model is used to minimize the Wasserstein distance between model distribution and data distribution, thereby extracting the intrinsic relationship between the mean vector and variance term. 

In the implementation process of DVNE, the $W_2$ distance of two Gaussian distribution functions is defined as:

\begin{equation}
W_2\left(N\left(\mu_1, \boldsymbol{\Sigma}_1\right) ; N\left(\mu_2, \boldsymbol{\Sigma}_2\right)\right)^2= \left\|\mu_1-\mu_2\right\|+\left\|\boldsymbol{\Sigma}_1^{\frac{1}{2}}-\boldsymbol{\Sigma}_2^{\frac{1}{2}}\right\|_{\mathrm{F}}^2
\end{equation}
where, N represents the Gaussian distribution. The loss function L of DVNE consists of two parts: one is the loss based on ranking that keeps the first order approximate L1 norm; The second is to maintain the second-order approximation L2 norm reconstruction loss.

\begin{equation}
    \begin{gathered}L=L_1+\alpha L_2, \\ L_1=\sum_{(i, j, k) \in D}\left(E_{i j}^2+\exp \left(-E_{i k}\right)\right), \\ L_2=\inf _{Q(Z \mid C) \in Q} E_{P_{\boldsymbol{C}}} E_{Q(Z \mid C)}\left[\|C \odot(\boldsymbol{C}-G(Z))\|_2^2\right]\end{gathered}
\end{equation}
where $D=\{(i, j, k) \mid j \in N(i), k \notin N(j)\}$ is a triple set, $E_{ij}$ is the $W_2$ distance between node $i$ and $j$. $C$ is the input feature, $Q$ is the encoder, $\bigodot$ is the Hadamard product, $G$ is the decoder, $Z$ is the random variable. Finally, the parameters in the model are learned by minimizing the loss function.

The method\cite{l34} introduced AE into the learning representation of vertices and proposed a structured deep network embedding (SDNE) method. Most existing network embedding methods use shallow models, which cannot capture highly nonlinear network structures, resulting in poor network performance. The SDNE method utilizes second-order approximation to capture global network structures, The network performance is not good enough. At the same time, the first-order approximation is used to maintain the local network structure. Finally, the network structure is maintained by using first-order and second-order proximity in the semi supervised depth model, which can effectively capture the highly nonlinear network structure and maintain the global and local structures. Then the loss function of the model is:
\begin{equation}
    L_{min}=L_{2nd}+\alpha L_{1st}+ vL_{reg}
\end{equation}
where L2 is the second order approximate loss function, L1 is the first order approximate loss function, and Lr is the regularization term to prevent overfitting. Each loss function is defined as:
\begin{equation}
    \begin{gathered}L_{2 \mathrm{nd}}=\sum_{i=1}^n\left\|\left(\hat{\boldsymbol{o}}_i-\boldsymbol{o}_i\right) \odot \boldsymbol{b}_i\right\|_2^2=  \|(\hat{\boldsymbol{X}}-\boldsymbol{X}) \odot \boldsymbol{B}\|_{\mathrm{F}}^2, \\ L_{1 \mathrm{st}}=\sum_{i, j=1}^n a_{i, j}\left\|\boldsymbol{h}_i-\boldsymbol{h}_j\right\|_2^2\end{gathered}
\end{equation}
where $\boldsymbol{b}_i=\left\{b_{i, j}\right\}_{j=1}^n$, if $a_{i,j}=0$, $b_{i,j}=1$, otherwise $b_{i,j}=\beta>1$, $b$ and $\beta$ are parameters. $a_{i,j}$ is the element of adjacency matrix $A$.

Rule equivalence refers to the fact that vertices located in different parts of a network may have similar roles or positions, which is easily overlooked in research on network embedding. Reference\cite{l35} proposes a deep recursive network embedding (DRNE) to learn network embeddings with rule equivalency. The neighborhood of nodes is transformed into an ordered sequence, and each node is represented by a normalized LSTM layer, Aggregates their neighbor characteristics by recursion, and the loss function of DRNE is:
\begin{equation}
    L = \sum_{v \in V} || X_v-Agg (\lbrace X_u|u \in N(v) \rbrace) ||_F^2  
\end{equation}
where, $X_v$ and $X_u$ represent the embedded vector representation of node $v$ and $u$, and $Agg$ is a aggregate function implemented by LSTM. In a recursive step, the embedded representation of nodes can maintain the local structure of their neighborhood. By iteratively updating the learned representation, the learned embedded vector of nodes can integrate their structural information in a global sense, so as to achieve rule equivalence. When serializing neighborhood nodes, The most effective neighborhood ranking measure - degree - is used to rank them. Finally, regularization term is added as the loss function of the whole model to update the parameters.

\begin{table}[]
\caption{Autoencoder-Based Graph Neural Network Models}
\centering
\setlength{\tabcolsep}{10pt}
\renewcommand{\arraystretch}{1.5}
\resizebox{\textwidth}{!}{
\begin{tabular}{ccccp{6cm}}
\hline
Model & Year & Dataset & Application & \thead{Disadvantage} \\
\hline
VGAE\cite{l32} & 2016 & \makecell{Citeseer, Cora,\\Pubmed} & Link Prediction & Limited expressiveness, struggles with higher-order dependencies \\
DVNE\cite{l33} & 2018 & \makecell{Cora, Facebook,\\BolgCatelog, Flickr} &Link prediction classification &  High complexity, scalability issues \\
SDNE\cite{l34} & 2016 & \makecell{Arive-Grqc, BlogCatalog} &Link prediction classification & Sensitive  hyperparameters, potential over-smoothing, limited structure capture \\
DRNE\cite{l35} & 2018 & \makecell{BlogCatalog, Jazz} & Classification & High memory usage, limited scalability \\
GC-MC\cite{l36} & 2017 & \makecell{MovieLens, Flixster,\\Douban, YahooMusic} & Matrix completion & Limited long-range dependency capture, potential generalization issues \\
ARGA/ARVGA\cite{l37} & 2018 & \makecell{Citeseer, Cora,\\Pubmed} & Link prediction graph clustering & Computational inefficiency, challenges in global structure modeling \\
Graph2Gauss\cite{l38} & 2018 & \makecell{Citeseer, Cora,\\Pubmed} & Link prediction classification & Limited complex structure representation, difficulty with dynamics \\
\hline
\end{tabular}
}
\end{table}

The method\cite{l36} applied AE to matrix completion in recommendation systems and proposed the graph convolution matrix completion method (GC-MC). GC-MC viewed matrix completion as a link prediction problem on the graph and designed a graph self coding framework based on bipartite interaction graph for differentiable information transmission. Its encoder was implemented using graph convolution, The decoder is completed by a bilinear function. Reference\cite{l37} proposes a new framework for combating graph embedding of graph data. In order to learn robust embedding, two countermeasures are proposed to combat regularization graph auto encoder (ARGA) and regularization variational graph auto encoder (ARVGA). In addition to the above methods, The graph neural network based on autoencoders also has the Graph2Gauss\cite{l38} method that can effectively learn node embeddings on large-scale graphs. Table 2 summarizes the graph neural network models based on autoencoders.

\section{\it\bf Experiments on Graph Classification and Link Prediction}
\subsection{GNN Classifier}
Let $\tilde{H}^1$ be the augmented node repersentation set by concatenating $\tilde{H}^1$ with the embedding of the synthetics nodes, and $\tilde{V}_L$ be the augmented labeled set by incorporating the synthetic nodes into $V_L$. We have an augmented graph $\tilde{G} = \{ \tilde{A},\tilde{H} \}$ with labeled node set $\tilde{V}_L$. The data size of different classes in  $\tilde{G}$ becomens balanced, and an unbiased Gnn classifier would be abel to be trained on that. Specifically, we adopt another GraphSage block, appended by a linear layer for node classification on $\tilde{G}$ as:
\begin{equation}
\mathbf{h}_v^2=\sigma\left(\mathbf{W}^2 \cdot \operatorname{CONCAT}\left(\mathbf{h}_v^1, \tilde{\mathbf{H}}^1 \cdot \tilde{\mathbf{A}}[:, v]\right)\right)
\end{equation}

\begin{equation}
\mathbf{P}_v=\operatorname{softmax}\left(\sigma\left(\mathbf{W}^c \cdot \operatorname{CONCAT}\left(\mathbf{h}_v^2, \mathbf{H}^2 \cdot \tilde{\mathbf{A}}[:, v]\right)\right)\right)
\end{equation}

where $H^2$ represents node representation matrix of the 2nd GraphSage block, and $W$ refers to the weight parameters. $P_v$ is the probability distribution on class labels for node v. The classifier module is optimized using cross-entropy loss as:

\begin{equation}
    \mathbf{Y}_v^{\prime}=\underset{c}{\operatorname{argmax}} \mathbf{P}_{v, c}
\end{equation}

We compare the some GNN-based models' performance. Table 3 shows the corresponding $F_1$ and $MCC$ values for the two real-world datasets.

\begin{table}[htbp]
\caption{Classification Performance on GNN-based Models}
\setlength{\tabcolsep}{5pt}
\begin{tabular}{llllllll}
\hline
Method               & Cora                   &                        &                      &  & CiteSeer               &                        &          \\ \cline{2-4} \cline{6-8} 
                     & F1                     & MCC                    & p-value              &  & F1                     & MCC                    & p-value  \\ \hline
GCN\cite{GCN}                  & 0.6861±0.0023          & 0.6146±0.0014          & 1.98e-19             &  & 0.6158±0.0029          & 0.5549±0.0014          & 9.24e-16 \\
GAT\cite{l21}                  & 0.7134±0.072           & 0.6379±0.0061          & 3.90e-09             &  & 0.6290±0.0085          & 0.5662±0.01807         & 6.78e-07 \\
GraphSMOTE\cite{GraphSMOTE}           & 0.7213±0.0075          & 0.6553±0.0066          & 1.19e-13             &  & 0.6294±0.0091          & 0.6113±0.0083          & 1.09e-05 \\
DR-GCN\cite{DR}               & 0.7247±0.0057          & 0.6588±0.0065          & 6.95e-09             &  & 0.6332±0.0049          & 0.6143±0.0038          & 2.74e-09 \\
GNN-INCM\cite{GNN-INCM}             & \textbf{0.7508±0.0045} & \textbf{0.7273±0.0051} & -                    &  & \textbf{0.6490±0.0048} & \textbf{0.6274±0.0036} & -        \\ \hline
\multicolumn{1}{c}{} & \multicolumn{1}{c}{}   & \multicolumn{1}{c}{}   & \multicolumn{1}{c}{} &  & \multicolumn{1}{c}{}   &                        &         
\end{tabular}
\end{table}

\subsection{Link Prediction}
Human trajectory prediction is one application of link prediction.Two metrics are used to evaluate model performance: the Average Displacement Error (ADE) \cite{liu10} defined in equation \ref{eq4} and the Final Displacement Error (FDE) \cite{liu11} defined in equation \ref{eq5}. Intuitively, ADE measures the average prediction performance along the trajectory, while the FDE considers only the prediction precision at the end points. Since Social-STGCNN generates a bi-variate Gaussian distribution as the prediction, to compare a distribution with a certain target value, we follow the evaluation method used in Social-LSTM \cite{liu11} in which 20 samples are generated based on the predicted distribution. Then the ADE and FDE are computed using the closest sample to the ground truth. This method of evaluation were adapted by several works such as Social-GAN \cite{liu12} and many more.
The performance of Social-STGCNN is compared with other models on ADE/FDE metrics in table 4 \cite{liu7}.

\begin{equation}
\mathrm{ADE}=\frac{\sum_{n \in N} \sum_{t \in T_p}\left\|\hat{p}_{\mathrm{t}}^n-p_{\mathrm{t}}^n\right\|_2}{N \times T_p} \label{eq4}
\end{equation}

\begin{equation}
\mathrm{FDE}=\frac{\sum_{n \in N}\left\|\hat{p}_{\mathrm{t}}^n-p_{\mathrm{t}}^n\right\|_2}{N} ,t = T_p\label{eq5}
\end{equation}

\begin{table}[htbp]
\setlength{\tabcolsep}{30pt}
\centering
\caption{ADE / FDE Metrics for Several Methods Compared to Social-STGCNN are Shown.}
\begin{tabular}{cccc}
\toprule 
 & ETH  & HOTEL & UNIV \\
\midrule 
Linear\cite{liu11}  & 1.33/2.94 & 0.39/0.72 & 0.82/1.59 \\
S-LSTM \cite{liu11} & 1.09/2.35 & 0.79/1.76 & 0.67/1.40 \\
CGNS \cite{liu13}   & \textbf{0.62}/1.40 & 0.70/0.93 & 0.48/1.22 \\
PIF \cite{liu14}    & 0.73/1.65 & \textbf{0.30/0.59} & 0.60/1.27 \\
Social-STGCNN \cite{st}      & 0.64/\textbf{1.11} & 0.49/ 0.85 & \textbf {0.44/0.79} \\
\bottomrule 
\end{tabular}
\end{table}

\renewcommand\refname{References}


\begin{thebibliography}{99}

\bibitem{l1}
LeCun Y, Bottou L, Bengio Y, et al. Gradient-based learning applied to document recognition[J]. Proceedings of the IEEE, 1998, 86(11): 2278-2324.

\bibitem{l2}
Gu J, Wang Z, Kuen J, et al. Recent advances in convolutional neural networks[J]. Pattern recognition, 2018, 77: 354-377.

\bibitem{l3}
Lawrence S, Giles C L, Tsoi A C, et al. Face recognition: A convolutional neural-network approach[J]. IEEE transactions on neural networks, 1997, 8(1): 98-113.

\bibitem{l4}
Ciresan D C, Meier U, Masci J, et al. Flexible, high performance convolutional neural networks for image classification[C]//Twenty-second international joint conference on artificial intelligence. 2011.

\bibitem{l5}
Dalal N, Triggs B. Histograms of oriented gradients for human detection[C]//2005 IEEE computer society conference on computer vision and pattern recognition (CVPR'05). Ieee, 2005, 1: 886-893.

\bibitem{l6}
Fan J, Xu W, Wu Y, et al. Human tracking using convolutional neural networks[J]. IEEE transactions on Neural Networks, 2010, 21(10): 1610-1623.

\bibitem{l7}
Huang W, Qiao Y, Tang X. Robust scene text detection with convolution neural network induced mser trees[C]//Computer Vision–ECCV 2014: 13th European Conference, Zurich, Switzerland, September 6-12, 2014, Proceedings, Part IV 13. Springer International Publishing, 2014: 497-511.

\bibitem{l8}
Chen Y. Convolutional neural network for sentence classification[D]. University of Waterloo, 2015.

\bibitem{l9}
Donahue J, Jia Y, Vinyals O, et al. Decaf: A deep convolutional activation feature for generic visual recognition[C]//International conference on machine learning. PMLR, 2014: 647-655.

\bibitem{l10}
Zhu H, Li X, Zhang P, et al. Learning tree-based deep model for recommender systems[C]//Proceedings of the 24th ACM SIGKDD International Conference on Knowledge Discovery \& Data Mining. 2018: 1079-1088.

\bibitem{l11}
Niepert M, Ahmed M, Kutzkov K. Learning convolutional neural networks for graphs[C]//International conference on machine learning. PMLR, 2016: 2014-2023.

\bibitem{l12}
Gao H, Wang Z, Ji S. Large-scale learnable graph convolutional networks[C]//Proceedings of the 24th ACM SIGKDD international conference on knowledge discovery \& data mining. 2018: 1416-1424.

\bibitem{l13}
Atwood J, Towsley D. Diffusion-convolutional neural networks[J]. Advances in neural information processing systems, 2016, 29.

\bibitem{l14}
Diao Z, Wang X, Zhang D, et al. Dynamic spatial-temporal graph convolutional neural networks for traffic forecasting[C]//Proceedings of the AAAI conference on artificial intelligence. 2019, 33(01): 890-897.

\bibitem{l15}
Qi S, Wang W, Jia B, et al. Learning human-object interactions by graph parsing neural networks[C]//Proceedings of the European conference on computer vision (ECCV). 2018: 401-417.

\bibitem{l16}
Xu B, Cen K, Huang J. A Survey on Graph Convolutional Neural Network[J]. Chinese Journal of Computers,2020,43(5):755-780. DOI:10.11897/SP.J.1016.2020.00755.

\bibitem{l17}
Wang J, Kong L, Huang Z, et al. Survey of Graph Neural Network[J]. Computer Engineering,2021,47(4):1-12. DOI:10.19678/j.issn.1000-3428.0058382.

\bibitem{l18}
Ma S, Liu J, Zuo X. Survey on Graph Neural Networ[J]. Journal of Computer Research and Development,2022,59(1):47-80. DOI:10.7544/issn1000-1239.20201055.

\bibitem{l19}
Perozzi B, Al-Rfou R, Skiena S. Deepwalk: Online learning of social representations[C]//Proceedings of the 20th ACM SIGKDD international conference on Knowledge discovery and data mining. 2014: 701-710.

\bibitem{l20}
Yang Z, Cohen W, Salakhudinov R. Revisiting semi-supervised learning with graph embeddings[C]//International conference on machine learning. PMLR, 2016: 40-48.

\bibitem{l21}
Veličković P, Cucurull G, Casanova A, et al. Graph attention networks[J]. arXiv preprint arXiv:1710.10903, 2017.

\bibitem{l22}
Kipf T N, Welling M. Semi-supervised classification with graph convolutional networks[J]. arXiv preprint arXiv:1609.02907, 2016.

\bibitem{l23}
Xu B, Shen H, Cao Q, et al. Graph wavelet neural network[J]. arXiv preprint arXiv:1904.07785, 2019.

\bibitem{liu4} 
Yan S, Xiong Y, Lin D. Spatial temporal graph convolutional networks for skeleton-based action recognition[C]//Thirty-second AAAI conference on artificial intelligence. 2018.

\bibitem{l24}  
Bahdanau D, Cho K, Bengio Y. Neural machine translation by jointly learning to align and translate[J]. arXiv preprint arXiv:1409.0473, 2014.

\bibitem{l25}
Zhang S, Xie L. Improving attention mechanism in graph neural networks via cardinality preservation[C]//IJCAI: Proceedings of the Conference. NIH Public Access, 2020, 2020: 1395.

\bibitem{l26}
Zhang J, Shi X, Xie J, et al. GaAn: Gated attention networks for learning on large and spatiotemporal graphs[J]. arXiv preprint arXiv:1803.07294, 2018.

\bibitem{l27}
Zhang R, Zou Y, Ma J. Hyper-SAGNN: a self-attention based graph neural network for hypergraphs[J]. arXiv preprint arXiv:1911.02613, 2019.

\bibitem{l28}
Nguyen D Q, Nguyen T D, Phung D. Universal graph transformer self-attention networks[C]//Companion Proceedings of the Web Conference 2022. 2022: 193-196.

\bibitem{l29}
Chang H, Rong Y, Xu T, et al. Spectral graph attention network with fast eigen-approximation[C]//Proceedings of the 30th ACM International Conference on Information \& Knowledge Management. 2021: 2905-2909.

\bibitem{l30}
Knyazev B, Taylor G W, Amer M. Understanding attention and generalization in graph neural networks[J]. Advances in neural information processing systems, 2019, 32.

\bibitem{l31}
Xu K, Hu W, Leskovec J, et al. How powerful are graph neural networks?[J]. arXiv preprint arXiv:1810.00826, 2018.

\bibitem{l32}
Kipf T N, Welling M. Variational graph auto-encoders[J]. arXiv preprint arXiv:1611.07308, 2016.

\bibitem{l33}
Zhu D, Cui P, Wang D, et al. Deep variational network embedding in wasserstein space[C]//Proceedings of the 24th ACM SIGKDD international conference on knowledge discovery \& data mining. 2018: 2827-2836.

\bibitem{l34}
Wang D, Cui P, Zhu W. Structural deep network embedding[C]//Proceedings of the 22nd ACM SIGKDD international conference on Knowledge discovery and data mining. 2016: 1225-1234.

\bibitem{l35}
Tu K, Cui P, Wang X, et al. Deep recursive network embedding with regular equivalence[C]//Proceedings of the 24th ACM SIGKDD international conference on knowledge discovery \& data mining. 2018: 2357-2366.

\bibitem{l36}
Berg R, Kipf T N, Welling M. Graph convolutional matrix completion[J]. arXiv preprint arXiv:1706.02263, 2017.

\bibitem{l37}
Pan S, Hu R, Long G, et al. Adversarially regularized graph autoencoder for graph embedding[J]. arXiv preprint arXiv:1802.04407, 2018.

\bibitem{l38}
Bojchevski A, Günnemann S. Deep gaussian embedding of graphs: Unsupervised inductive learning via ranking[J]. arXiv preprint arXiv:1707.03815, 2017.

\bibitem{liu8} 
Zhang M, Cui Z, Neumann M, et al. An end-to-end deep learning architecture for graph classification[C]//Proceedings of the AAAI conference on artificial intelligence. 2018, 32(1).

\bibitem{liu9} 
Kipf T N, Welling M. Semi-supervised classification with graph convolutional networks[J]. arXiv preprint arXiv:1609.02907, 2016.

\bibitem{liu10} 
Alahi A, Goel K, Ramanathan V, et al. Social lstm: Human trajectory prediction in crowded spaces[C]//Proceedings of the IEEE conference on computer vision and pattern recognition. 2016: 961-971.

\bibitem{liu11} 
Pellegrini S, Ess A, Schindler K, et al. You'll never walk alone: Modeling social behavior for multi-target tracking[C]//2009 IEEE 12th international conference on computer vision. IEEE, 2009: 261-268.

\bibitem{liu12} 
Gupta A, Johnson J, Fei-Fei L, et al. Social gan: Socially acceptable trajectories with generative adversarial networks[C]//Proceedings of the IEEE conference on computer vision and pattern recognition. 2018: 2255-2264.

\bibitem{liu13} 
Li J, Ma H, Tomizuka M. Conditional generative neural system for probabilistic trajectory prediction[C]//2019 IEEE/RSJ International Conference on Intelligent Robots and Systems (IROS). IEEE, 2019: 6150-6156.

\bibitem{liu14} 
Liang J, Jiang L, Niebles J C, et al. Peeking into the future: Predicting future person activities and locations in videos[C]//Proceedings of the IEEE/CVF conference on computer vision and pattern recognition. 2019: 5725-5734.

\bibitem{st}
Mohamed A, Qian K, Elhoseiny M, et al. Social-stgcnn: A social spatio-temporal graph convolutional neural network for human trajectory prediction[C]//Proceedings of the IEEE/CVF conference on computer vision and pattern recognition. 2020: 14424-14432.

\bibitem{liu7} 
Mohamed A, Qian K, Elhoseiny M, et al. Social-stgcnn: A social spatio-temporal graph convolutional neural network for human trajectory prediction[C]//Proceedings of the IEEE/CVF conference on computer vision and pattern recognition. 2020: 14424-14432.

\bibitem{GCN}
Kipf T N, Welling M. Semi-supervised classification with graph convolutional networks[J]. arXiv preprint arXiv:1609.02907, 2016.

\bibitem{DR}
Shi M, Tang Y, Zhu X, et al. Multi-class imbalanced graph convolutional network learning[C]//Proceedings of the Twenty-Ninth International Joint Conference on Artificial Intelligence (IJCAI-20). 2020.

\bibitem{GraphSMOTE}
Zhao T, Zhang X, Wang S. Graphsmote: Imbalanced node classification on graphs with graph neural networks[C]//Proceedings of the 14th ACM international conference on web search and data mining. 2021: 833-841.

\bibitem{GNN-INCM}
Huang Z, Tang Y, Chen Y. A graph neural network-based node classification model on class-imbalanced graph data[J]. Knowledge-Based Systems, 2022, 244: 108538.

\end{thebibliography}
\end{document}